\documentclass[12pt]{article}

\usepackage{amsmath,amssymb}
\usepackage{latexsym}
\usepackage{bm}
\usepackage{graphicx}

\newcommand{\bee}{\begin{equation*}}
\newcommand{\eee}{\end{equation*}}
\newcommand{\be}{\begin{equation}}
\newcommand{\ee}{\end{equation}}

\begin{document}

\markboth{A. G. Ramm}{      }

\title{Dimension reduction in representation of the data
}

\author{A. G. Ramm$\dag$\\
\\
$\dag$Mathematics Department, Kansas State University,\\
Manhattan, KS 66506-2602, USA\\
email: ramm@math.ksu.edu\\
}

\date{}
\maketitle

\begin{abstract} \noindent 
Suppose the data consist of a set $S$ of points $x_j$, $1\leq j \leq J$, 
distributed in a bounded domain $D\subset R^N$, where $N$ is a large 
number. An algorithm is given for finding the sets $L_k$ of dimension 
$k\ll N$, $k=1,2,...K$,
in a neighborhood of which maximal amount of points $x_j\in S$ lie.
The algorithm is different from PCA (principal component analysis).

  \end{abstract}




\section{Statement of the problem and the description 
of the algorithm
\label{s1}}

In many applications the data are presented as a set $S$ of points $x_j$,
$1\leq j \leq J$, $x_j\in D\subset R^N$, where $J$ is a very large number,
$D$ is a known bounded domain, for example, a box, and $N$ is a large 
number. It is useful practically to have a more economical data 
representation, if this is possible. For instance, there may be a case
when the data points are concentrated in a neighborhood of some set
$L$ of dimension $k\ll N$. In this case one would like to find
this set. This problem is an old one. One widely known version of it
is the regression problem. In its simplest formulation the regression 
problem consists of finding a straight line $y=a_1x+a_2$ which
represents the set of data points $\{\xi_j, \eta_j\}_{j=1}^J$, in $R^2$
optimally in the sense $\sum_{j=1}^J(a_1\xi_j+a_2-\eta_j)^2=\min$, where
the minimization is taken with respect to $a_1$ and $a_2$. This problem is
well studied in statistics. Analogous formulations can be done
under the assumption that the regression curve is not a straight line but 
some function, depending on finitely many parameters $a_m$, $1\leq m \leq 
M$. A different approach to the problem of the dimension reduction
in the representation of the data was proposed in 1901 by K.Pearson,
in a paper entitled "On lines and planes of closest fit to systems of 
points in space". This paper and many subsequent papers in which the 
theory of PCA (principal components analysis) was developed are referenced 
in [G], where one can find the very recent survey papers on the problem
of dimension reduction in representation of the data. The PCA theory
in its simplest version which preassumes that the data points in $R^2$ are 
concentrated in a neighborhood of a straight line $L$, consists of
finding $L$ from the minimization problem: 
$\sum_{j=1}^Jd_j^2=\min,$ where $d_j$ is the distance from the point
$\{\xi_j, \eta_j\}$ to the straight line $L$. The minimization
is taken with resepct to parameters which define the straight line $L$, 
for example, with respect to $a_1$ and $a_2$. There is a difference 
between the regression problem and the PCA problem: in the regression 
problem one minimizes not the sum of the squares of the distances from the 
points $\{\xi_j, \eta_j\}$ to $L$, but the sum of the squares of
the lengths of the vertical segments from $\{\xi_j, \eta_j\}$ to $L$.
A priori it is not known if a straight line is the set in a neighborhood
of which most of the points of $S$ lie. 

The aim of this paper is to propose an algorithm for computing the 
set $L_k$ of dimension $k\ll N$ in a neighborhood of which many points of 
$S$ lie. The set $L_k$ that we construct, is a polyhedron with vertices in 
an $r$-neighborhood of which many points of $S$ lie. By an 
$r$-neighborhood of a pont $y\in R^N$ the ball $B(y,r):=\{x: |x-y|\leq r, 
x\in R^N$ is meant, $|x-y|$ is the Euclidean distance between points $x$ 
and $y$ in $R^N$.

Our algorithm does not preassume that the clusters of the points should
lie near a linear manifold or near a non-linear manifold which is
a priori known up to a finitely many parameters.

Let us now decsribe the steps of our algorithm for computing the set $L$ 
in an $r$-neighborhood of which many points of $S$ lie.

1. Fix a number $r>0$ and a cubic grid with the step-size $r$ in $R^N$.
Let $y_m$ be the nodes of this grid, $1\leq m \leq M$, and $B_m$ be the 
ball of radius $r$ centered at $y_m$. 

2. Scan the domain $D$, in which the set $S$ of the data points $x_j$
lies, by moving the ball $B_m$ so that $m$ runs from $1$ to $M$, that is,
the center of the ball runs through all the nodes of the grid  
belonging to $D$. Each of the points of $S$ will belong to some ball 
$B_m$. Calculate the number $\nu_m$ of the points of $S$ in $B_m$, and 
arrange the numbers $\nu_m$ in a descending order: $\nu_1\geq \nu_2\geq 
\nu_3......$. Let $y_k$ be the center of the ball $B_m$ containing
$\nu_k$ points. Fix some threshold number $\nu$ and  
neglect the balls containing less than $\nu$ points. Let $K=K(\nu)$
be the number such that $\nu_k>\nu$ for $k\leq K$ and $\nu_k\leq \nu$
for $k>K$.      

3. Define $L^{1}$ to be the one-dimensional set of segments, joining $y_k$
and $y_{k+1}$. Then  $L^{1}$ is a one-dimensional set, a union of segments
in $R^N$, and in $r-$neighborhood of the vertices of this set, i.e., 
of the points $y_k$, $1\leq k \leq K$, one has many points of the set $S$.
There is no guarantee that there are points of $S$ near every point of the 
set $L^{1}$. 

One may change the algorithm by choosing the nearest to $y_1:=z_1$
point among the points $\{y_k\}_{y_k\neq y_1}$, denoting this point $z_2$,
and then choosing the closest to $z_2$ point $z_3$ among the points 
$\{y_k\}_{y_k\neq z_1,\, y_k\neq z_2}$, and continuing in this fashion one
gets the set of points $z_k$, $1\leq k \leq K$. Joining  $z_k$
and  $z_{k+1}$ by a segment and denoting $L^{1}_{z}$ the union of these
segments, one gets a one-dimensional set of points such that
in an $r-$neighborhood of its vertices there are many points of $S$.
In such a way one may construct more than one  line: it might happen that two 
(or more) intersecting or non-intersecting lines will be constructed.

One may consider the triangles $T_k$ with vertices $z_k, z_{k+1}, 
z_{k+2}$, $1\leq k \leq K-2$. The union of  $T_k$ forms a 
two-dimensional set in $R^N$. In an $r-$neighborhood  
of its vertices there are many points of $S$.

One may construct in a similar way the sets of dimension
$s$ in  $R^N$, such that in an $r-$neighborhood of its vertices
there are many points of $S$.  

The threshold number $\nu$ is not known a priori, and one starts,
e.g., with $\nu=10^3$, and if there are few balls with $\nu_k>\nu$,
then one may restart the procedure with $\nu=10^2$. If, on the other hand,
there are very many balls with $\nu_k>\nu$, then one may restart
the procedure with $\nu=10^4$. Also, the parameter $r$ may be 
treated similarly.



\begin{thebibliography}{99}
\small
\bibitem{G} A.N.Gorban, B.Kegl, D.Wunsch, A.Zinovyev, (Editors), 
{\it Principal manifolds for data 
visualization and dimension reduction}, Lecture notes in 
computational science and engineering, Vol. 58, 
Springer, Berlin, (2008).





\end{thebibliography}
\end{document}